\documentclass[preprint,12pt]{elsarticle}

\usepackage{amssymb}

\usepackage{amsmath} % 数学符号

\usepackage{algorithm}
\usepackage{algpseudocode}
\usepackage{amsmath}
\usepackage{subfloat}%插入并排图
\usepackage{float}
\usepackage{subfig}

\pdfoutput=1

\begin{document}
\begin{frontmatter}

\title{Local Structure-aware Graph Contrastive Representation Learning}

\author[inst1]{Kai Yang}
\author[inst1]{Yuan Liu}
\author[inst2]{Zijuan Zhao}
\author[inst1]{Peijin Ding}
\author[inst1]{Wenqian Zhao}

\affiliation[inst1]{organization={College of Information Engineering, Yangzhou University},%Department and Organization
            % addressline={Address One}, 
            city={Yangzhou},
            postcode={225127}, 
            % state={State One},
            country={China}}

\affiliation[inst2]{organization={Business School, University of Shanghai for Science and Technology},%Department and Organization
            % addressline={Address Two}, 
            city={Shanghai},
            postcode={200093}, 
            % state={State Two},
            country={China}}

\begin{abstract}
%% Text of abstract   \textbf{
Traditional Graph Neural Network (GNN), as a graph representation learning method, is constrained by label information. However, Graph Contrastive Learning (GCL) methods, which tackle the label problem effectively, mainly focus on the feature information of the global graph or small subgraph structure (e.g., the first-order neighborhood). In the paper, we propose a \textbf{L}ocal \textbf{S}tructure-aware \textbf{G}raph \textbf{C}ontrastive representation \textbf{L}earning method (LSGCL) to model the structural information of nodes from multiple views. Specifically, we construct the semantic subgraphs that are not limited to the first-order neighbors. For the local view, the semantic subgraph of each target node is input into a shared GNN encoder to obtain the target node embeddings at the subgraph-level. Then, we use a pooling function to generate the subgraph-level graph embeddings. For the global view, considering the original graph preserves indispensable semantic information of nodes, we leverage the shared GNN encoder to learn the target node embeddings at the global graph-level. The proposed LS-GCL model is optimized to maximize the common information among similar instances at three various perspectives through a multi-level contrastive loss function. Experimental results on five datasets illustrate that our method outperforms state-of-the-art graph representation learning approaches for both node classification and link prediction tasks.
\end{abstract}

%\centerline{
%\includegraphics[width=1 \linewidth]{lsgcl.eps}}

%\item We propose a novel contrastive learning framework named LS-GCL which utilizes the semantic subgraphs to extract particular local structural information at multiple perspectives.
%\item A multi-level contrastive loss function is provided to maximize common information among similar instances from three various perspectives including the node embeddings at the global graph-level and subgraph-level, subgraph-level graph embeddings corresponding to target nodes.
%\item Experimental results on five real-world datasets
%illustrate that the proposed LS-GCL model outperforms state-of-the-art graph representation learning approaches for both node classification and link prediction tasks.


\end{frontmatter}

\section{Introduction}

Graph neural network (GNN)\cite{fan2019graph, asif2021graph, wu2020comprehensive} has made noteworthy progress in processing the graph-structured data.
The core idea of GNN models, such as the graph convolutional network (GCN) \cite{gcn} and graph attention network (GAT) \cite{gat}, is to extract the structural information of nodes by transmitting and aggregating the attribute information of neighbors, and generate the embedding representations of nodes in the low-dimensional embedding space. 
The GNNs have widespread applications in real life, including the recommendation system \cite{wu2022graph}, intelligent transportation system \cite{zhou2020variational} and drug development in the biomedical field \cite{nguyen2021graphdta}, etc. 
However, traditional GNN methods, belonging to (semi-)supervised learning \cite{chen2021gnn, song2022graph}, are constrained by the label information, which is difficult to acquire in practical applications due to the privacy protection \cite{wu2022trustworthy}. The uneven distribution of labels or the false labels also affect the realistic performance of the GNN models \cite{yin2022generic}.

The unlabeled graph data (e.g., the adjacency matrix) is accessible compared with manual label information in real world. Graph contrastive learning (GCL) \cite{tao2022exploring}, as a popular self-supervised learning technique which utilizes known graph data to mine feature information of nodes or graphs, has attracted extensive researches. The core idea of GCL is to set the positive and negative sample objects according to the original graph information, and maximize the feature information between the target nodes and positive samples, minimize the common feature information between the target nodes and negative samples in the training process\cite{jin2020self}. Nodes with similar properties in the graph will generate close embedding representations in the embedding space. Recent GCL works have achieved excellent results in various downstream tasks, such as node classification \cite{nc}, graph classification \cite{gc} and link prediction \cite{lk}.
According to the contrastive objectives, existing GCL models can be summarized into three categories. 
The first is the Node-Graph level contrastive learning framework, which maximizes the common information between the target node embeddings (local graph structure) and the graph embedding of the global graph (global graph structure), such as the DGI\cite{velivckovic2018deep}.
The second is the Node-Node level contrastive learning framework, which maximizes the consistency of feature information between the target node embeddings (local graph structure) and the positive sample node embeddings (local graph structure), such as the GRACE\cite{zhu2020deep}, GCA\cite{zhu2021graph}, GMI\cite{peng2020graph}. 
The third is the Node-Subgraph level contrastive learning framework, which maximizes the common information between the target node embeddings (local graph structure) and the subgraph-level graph embedding (local graph structure), such as MVGRL\cite{hassani2020contrastive}, SUBG-CON\cite{jiao2020sub}. 
Among the three frameworks mentioned above, existing GCL models mainly focus on the global graph or small subgraph structure (e.g., the first-order neighborhood).
% The node embeddings cannot clearly capture their specific local structural and semantic information.

How to maximize the local feature information of nodes while preserving the global features of the original graph is a challenge.
In the paper, we propose a novel graph contrastive learning framework named \textbf{L}ocal \textbf{S}tructure-aware \textbf{G}raph \textbf{C}ontrastive representation \textbf{L}earning (LS-GCL) to model the feature information of nodes from multiple views. In the local perspective, the proposed LS-GCL constructs semantic subgraphs through selecting the relevant nodes of target nodes, and uses an encoder (GNN) to capture the structural information of semantic subgraphs for obtaining target node embeddings at the subgraph-level.
Moreover, to extract the local structural features of nodes more comprehensively,
we apply the mean-pooling function to generate the subgraph-level graph embeddings for each subgraph, which contain distinct local semantic information.
Considering the original graph preserves indispensable global information of target nodes, the LS-GCL leverages the shared GNN encoder to obtain the target node embeddings at global graph-level.
Finally, to integrate the local and global feature information, the LS-GCL proposes a multi-level contrastive loss function to maximize the common information among similar instances at three levels via a multi-level contrastive loss function. 
We conduct the node classification and link prediction experiments on five real-world datasets to validate the performance of the proposed LS-GCL. The experimental results demonstrate that the proposed LS-GCL achieves state-of-the-art results compared to classical GNN models and GCL methods. 
The source code of LS-GCL is available in https://github.com/LibertyAL/LS-GCL.

The principle contributions of the paper are as follows:

\begin{itemize}
\item[$\bullet$] We propose a novel contrastive learning framework named LS-GCL which utilizes the semantic subgraphs to extract particular local structural information at multiple perspectives.
\end{itemize}

\begin{itemize}
\item[$\bullet$] A multi-level contrastive loss function is provided to maximize common information among similar instances from three perspectives including the node embeddings at the global graph-level and subgraph-level, subgraph-level graph embeddings corresponding to target nodes.

\end{itemize}

\begin{itemize}
\item[$\bullet$] Experimental results on five real-world datasets
illustrate that the proposed LS-GCL model outperforms state-of-the-art graph representation learning approaches for both node classification and link prediction tasks.
\end{itemize}

\begin{figure}[!htb]%强制固定位置
\centerline{
\includegraphics[width=0.9\linewidth]{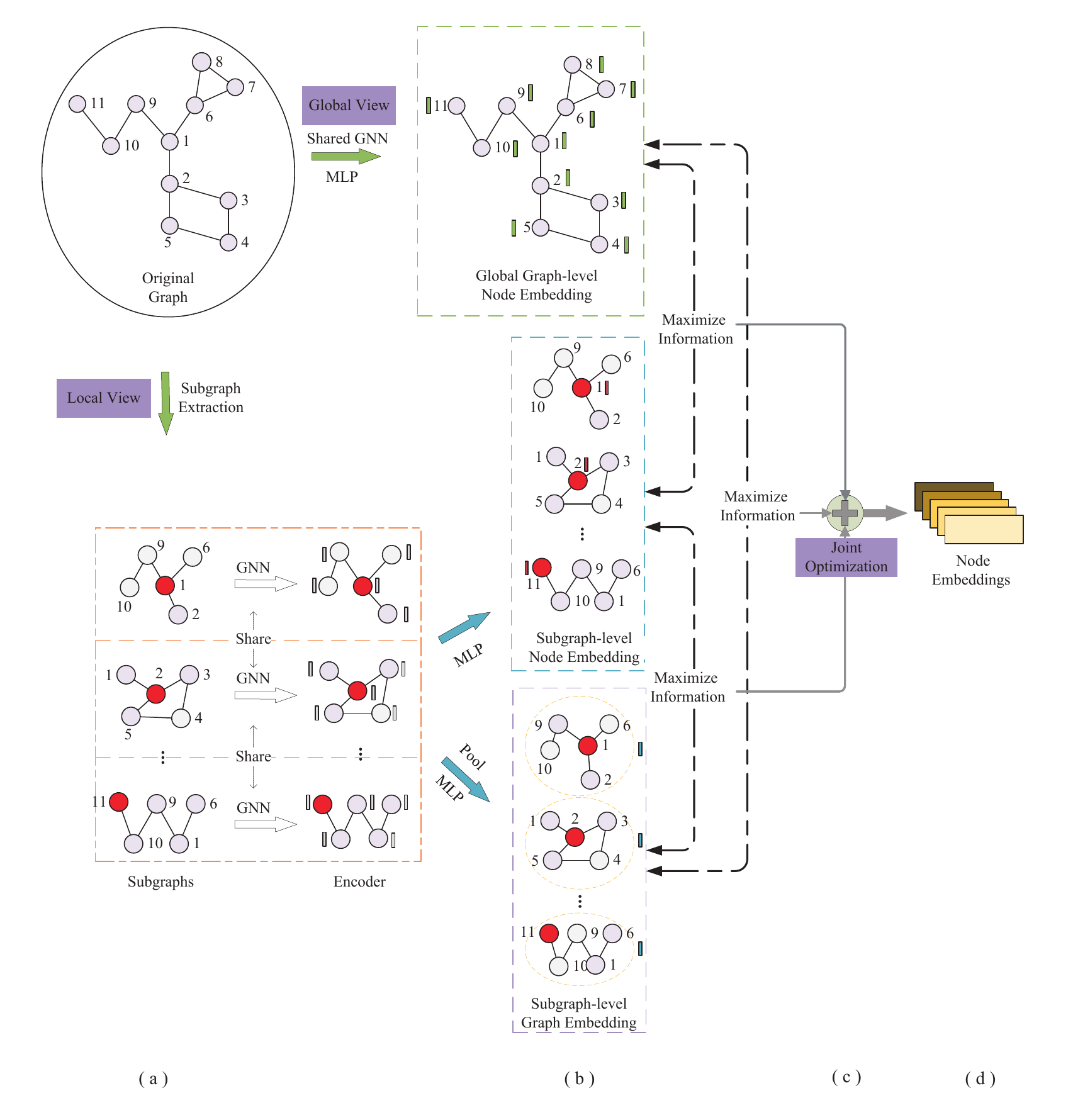}}
\caption{Framework of the proposed LS-GCL model. (a) Extract semantic subgraphs for each target node by the PPR algorithm. (b) Learn target node embeddings at global graph-level and subgraph-level, subgraph-level graph embeddings corresponding to target nodes. (c) Maximize the common information among similar instances at three levels via a multi-level contrastive loss function. (d) Generate the target node embeddings.
}
\label{fig:LSGCL}
\end{figure}

\section{Materials and Models}
In the section, we introduce the \textbf{L}ocal \textbf{S}tructure-aware \textbf{G}raph \textbf{C}ontrastive representation \textbf{L}earning method (LS-GCL). The architecture of our LS-GCL is shown in the Fig.\ref{fig:LSGCL}.
\subsection{Notations}
% 一些下面公式需要用到的符号可以再添加
Let $\mathcal{G} = (\mathcal{V}, \mathcal{E})$ indicates an undirected attributed graph, where $\mathcal{V}$ and $\mathcal{E}$ denote the set of nodes and edges, respectively. We denote the adjacency matrix of the graph and the initial feature matrix of the nodes as $\textbf{A} \in \left\{0,1 \right\}^{N \times N} $ and $\textbf{X} \in \mathbb{R}^{N \times F}$, where \emph{N} and \emph{F} represent the number of nodes in the graph and the dimension of the initial attribute feature, respectively. $\textbf{A}_{ij} = 1$ if ($i$, $j$) $\in \mathcal{E}$. $\textbf{X}_{i} \in \mathbb{R}^{1 \times F}$ denotes the initial feature of node $i$. 

\subsection{Data Augmentation with Semantic Subgraph}

In the GCL framework, the positive and negative samples for graph-structured data are constructed in terms of the graph structure or the attribute features of nodes, which affect the learning ability of the encoder\cite{feng2022adversarial}.
The DGI obtains the corrupted graph by shuffling the node order of the feature matrix and utilizes the corrupted graph embedding and original graph embedding as negative sample and positive sample separately \cite{velivckovic2018deep}.
The GGD\cite{zheng2022rethinking} obtains the corrupted graph by the edge and feature dropout, and assigns different manual labels to the nodes in the original graph and corrupted graph to train the GNN encoder\cite{zheng2022rethinking}.
However, global graph embedding, which compresses too much structural information into a simple fixed-length vector, approximates the constant vector\cite{zheng2022rethinking}, and the edge dropout methods ignore the information of the substructure.

In this paper, to explore the local graph structural information, we apply the semantic subgraph for data augmentation.
Considering the structural information of nodes mainly depends on local neighbors, we introduce the personalized pagerank algorithm (PPR) \cite{wang2020personalized}, which searches for the top $K$ related nodes to target nodes, to construct the semantic subgraphs.

Given the adjacency matrix $\textbf{A}$ of original graph $\mathcal{G}$, we firstly obtain the importance score matrix $\textbf{M}$, which is calculated as shown in Eq.\ref{eq:h2}:

\begin{equation}  %  align 多行公式 ,equation 单行公式
    \textbf{M} = p \cdot [\textbf{I} - (1-p) \cdot \hat{\textbf{A}}],
    \label{eq:h2}
\end{equation}
where $p$ is an optional parameter which is set as 0.15 and \textbf{I} denotes the identity matrix. $\hat{\textbf{A}} = \textbf{A} \textbf{D}^{-1}$ represents the column-normalized adjacency
matrix, where $\textbf{D}$ indicates the diagonal matrix with $\textbf{D}_{ii}=\sum_{j}\textbf{A}_{ij}$. And $\textbf{M}_{i}$ denotes the importance score vector of node $i$, where each value represents the relative importance between node $i$ and another node. 

For the target node $i$, we obtain the top $K$ important node set $ID(i)$ via the importance score vector $\textbf{M}_{i}$ of node $i$. The calculation process is shown below:

\begin{equation}  %  align 多行公式 ,equation 单行公式
    ID(i) = Rank(\textbf{M}_{i},K),
    \label{eq:h3}
\end{equation}
where $K$ indicates the count of nodes in the semantic subgraph and $Rank(\cdot)$ is a function that returns the index of the top $K$ related nodes. We construct semantic subgraphs via the obtained $K$ nodes and the relationships between them:
\begin{equation}  %  align 多行公式 ,equation 单行公式
    S_i=(\textbf{A}^{S}_i, \textbf{X}^{S}_i),
    \label{eq:h4}
\end{equation}
where $\textbf{X}^{S}_{i}$ and $\textbf{A}^{S}_{i}$ denote the attribute information and structural information of the semantic subgraph $S_i$ for target node $i$, respectively. The process of generating subgraphs can be achieved as a pre-task to reduce memory usage and running time compared to the online subgraph generation methods. 

\subsection{The contrastive objective} 

In the LS-GCL, each contrastive objective represents the local structural and semantic information of the target nodes in different levels. 
In this paper, we use a single GCN layer as the encoder to learn the node embeddings, and add a shared MLP layer for enhancing the expressiveness of the GNN encoder. The calculation process is as follows:

\begin{equation}  %  align 多行公式 ,equation 单行公式
    \textbf{H} = MLP \left[\sigma(\tilde{\textbf{D}}^{-1/2} \cdot \tilde{\textbf{A}}\cdot \tilde{\textbf{{D}}}^{-1/2} \cdot \textbf{X} \cdot \textbf{W}) \right],
    \label{eq:h1}
\end{equation}
where $\tilde{\textbf{A}}=\textbf{A}+\textbf{I}_{N}$ and $\textbf{I}_{N}$ denotes the identity matrix. $\tilde{\textbf{D}}_{ii}=\sum_{j}\tilde{\textbf{A}}_{ij}$ and \textbf{W} represents the weight matrix to be trained in the GNN encoder. $\sigma(\cdot)$ denotes the activation function (PReLU).
In our LS-GCL framework, the GNN encoder can be changed, such as the GAT, GraphSAGE and SGC which have different message aggregation methods.

% 通过一个编码器学习嵌入表示

For the semantic subgraph $S_i$ of target node $i$, we employ the GNN-based encoder to learn the target node embeddings. The calculation process is as follows:

\begin{equation}  %  align 多行公式 ,equation 单行公式
    \vec{h}^{S}_i = Encoder(\textbf{A}^{S}_i, \textbf{X}^{S}_i),
% \textbf{H}_i = GNN(\textbf{S}_i),
\label{eq:h5}
\end{equation}
where $\vec{h}^{S}_i$ represents the embedding of target node $i$ at the subgraph-level. The purpose of constructing semantic subgraph $S_i$ is to maximize the local structural information of nodes. Therefore, we apply a pooling function to generate the subgraph-level graph embedding for target node $i$:

\begin{equation}  %  align 多行公式 ,equation 单行公式
    \vec{h}^{i}_S = Pool(\textbf{H}_i),
    \label{eq:h7}
\end{equation}
where $Pool(\cdot)$ is the pooling function and $\vec{h}^{i}_S$ represents the embedding representation of the semantic subgraph $S_i$ for target node $i$.
In our experiments, we apply the mean-pooling function to learn the subgraph embeddings, which averages the feature vectors of nodes in the subgraph $S_i$.

Furthermore, we add the target node embeddings from the global-graph view as the contrastive objective.

\begin{equation}  %  align 多行公式 ,equation 单行公式
    \vec{h}^{G}_i = Encoder(\textbf{A}, \textbf{X}),
    \label{eq:h8}
\end{equation}
where $\vec{h}^{G}_i$ represents the embedding representation of target node $i$ at global graph-level.

Finally, we obtain the structural information of target nodes from multiple perspectives, including the node embedding at the global graph-level, the node embedding at the subgraph-level and the corresponding subgraph-level graph embedding, which can explain the semantic information of nodes from different levels.

\subsection{Multi-level Contrastive Learning Framework}

% 自监督学习的核心思想
In the paper, the process of LS-GCL is to define a pre-task that constructs positive and negative samples for training the GNN encoder without using label information. 
For the target node $i$, the LS-GCL combines the node embedding at the global graph-level $\vec{h}^{G}_i$ and the node embedding at subgraph-level $\vec{h}^{S}_i$ as well as the subgraph-level graph embedding $\vec{h}^{i}_S$. 
The LS-GCL then adopts the margin triplet loss function\cite{ha2021deep} to train the GNN encoder. To reduce the training time of the LS-GCL model, we conduct contrastive learning for any two perspectives in parallel. 
The contrastive loss function $\mathcal{L}_{NS}$ between the node embeddings $\vec{h}^{S}_i$ at subgraph-level and the subgraph-level graph embeddings $\vec{h}^{i}_S$ for target node $i$ is as follows:

\begin{equation}  %  align 多行公式 ,equation 单行公式
    \mathcal{L}_{NS}=\frac{1}{N}\sum_{i=0}^{N-1}(max(\sigma(\vec{h}^{S}_i \cdot \vec{h}^{i}_S)-\sigma(\vec{h}^{S}_i \cdot {\vec{h}}^{j}_S))+\alpha, 0),
    \label{eq:h12}
\end{equation}
where $\sigma(\cdot)$ denotes the sigmoid function. $\alpha$ is the margin value. ${\vec{h}}^{j}_S$ is the negative sample embedding where $j$ is one of the other nodes.

The contrastive loss function $\mathcal{L}_{NG}$ between the node embeddings $\vec{h}^{S}_i$ at subgraph-level and the node embeddings $\vec{h}^{G}_i$ at global graph-level for target node $i$ is as follows:

\begin{equation}  %  align 多行公式 ,equation 单行公式
    \mathcal{L}_{NG}=\frac{1}{N}\sum_{i=0}^{N-1}(max(\sigma(\vec{h}^{S}_i \cdot \vec{h}^{G}_i)-\sigma(\vec{h}^{S}_i \cdot {\vec{h}}^{G}_j))+\alpha, 0),
    \label{eq:h13}
\end{equation}

The contrastive loss function $\mathcal{L}_{SG}$ between the node embeddings $\vec{h}^{G}_i$ at the global graph-level and subgraph-level graph embeddings $\vec{h}^{i}_S$ for target node $i$ is as follows:

\begin{equation}  %  align 多行公式 ,equation 单行公式
    \mathcal{L}_{SG}=\frac{1}{N}\sum_{i=0}^{N-1}(max(\sigma(\vec{h}^{G}_i \cdot \vec{h}^{i}_S)-\sigma(\vec{h}^{G}_i \cdot {\vec{h}}^{j}_S))+\alpha, 0),
    \label{eq:h14}
\end{equation}

To capture the comprehensive structural and semantic information, the overall contrastive loss function $\mathcal{L}$ of the proposed LS-GCL can be defined as follows:

\begin{equation}  %  align 多行公式 ,equation 单行公式
    \mathcal{L}=\frac{1}{3}(\mathcal{L}_{NS} + \mathcal{L}_{NG} + \mathcal{L}_{SG}).
    \label{eq:h15}
\end{equation}

We adopt the Adam optimizer to adjust the parameters of the GNN encoder through the back propagation mechanism for maximizing the common information of node embeddings at three levels.
The pseudocode of the LS-GCL model is summarized in Algorithm \ref{alg:Framwork}.

\begin{algorithm}[htb]
\caption{Optimization Algorithm of the LS-GCL}
\label{alg:Framwork}
\begin{algorithmic}[1]
\Require
The global graph $\mathcal{G} = (\textbf{X}, \textbf{A})$;
The subgraphs for nodes $S_i = (\textbf{X}^{S}_i, \textbf{A}^{S}_i)$, $i=1,2, \ldots,N$;
% embedding dimension $d$;
\Ensure
Trained GNN-based encoder;
\State Extract semantic subgraphs for the target nodes;

\label{code:fram:extract}
\State Encode subgraphs $S_i$ and global graph $\mathcal{G}$ to obtain the node embedding $\vec{h}^{S}_i$ at the subgraph-level, $\vec{h}^{G}_i$ at the global graph-level and the graph embeddings $\vec{h}^{i}_S$ at subgraph-level;
\label{code:fram:add}
\State Compute the joint loss function $\mathcal{L}$ and update parameters of the GNN-based encoder;
\label{code:fram:classify}\\
\Return Trained GNN-based encoder;
\end{algorithmic}
\end{algorithm}

\section{Experiments and Results}
% 内容介绍:
In this section, we evaluate the effectiveness and robustness of the proposed LS-GCL model by the performances of node classification and link prediction experiments on five real-world datasets, and analyse the different components of the LS-GCL framework to identify their necessity.
% 数据集
\subsection{Datasets}
In the experiments, we utilize five public datasets, such as Cora\cite{CHEN2023505}, Citeseer\cite{TIAN2022294}, Pubmed\cite{ZHOU2023}, Cora\_ML\cite{geisler2020reliable} and DBLP\cite{tong2020directed} datasets. 
The five datasets are the citation graphs, where nodes represent scientific papers and edges represent the citation relationships between papers. Each node in the datasets has the initial features and a unique category label. 
The statistical information of the five datasets is shown in Table \ref{tab:table1}.

\begin{table}[htbp]
\renewcommand{\arraystretch}{1.3}
\setlength{\abovecaptionskip}{0pt}%    
\setlength{\belowcaptionskip}{10pt}%
\caption {The Statistical Information of the Five Datasets.}
\label{tab:table1}
\centering
\begin{tabular}{cccccc}
\hline
Datasets & Nodes & Edges  & Features & Classes \\ \hline
Cora     & 2708  & 5429  & 1433     &  7       \\
Cora\_ML & 2995  & 8158  & 2876     & 7       \\
Citeseer & 3327  & 4732   & 3703     &  6       \\
DBLP     & 17716 & 52867 & 1639     & 4       \\
Pubmed   & 19717 & 44338  & 500      & 3       \\ \hline
\end{tabular}
\end{table}

% 对比算法
\subsection{Baselines}
For the tasks of node classification and link prediction, we use two classical semi-supervised learning algorithms and four state-of-art graph contrastive learning algorithms as baseline algorithms. For the baseline methods, we report their experimental results on downstream tasks based on their open codes.
% to show the excellent performance of the proposed LS-GCL model in downstream tasks.
\begin{itemize}
\item \textbf{GCN}: The graph convolutional network (GCN) is a classical graph neural network algorithm that captures the structural information of target nodes by aggregating the feature information of neighbors to learn the node embeddings. The training process requires the participation of label information.
\item \textbf{GAT}: The graph attention network (GAT) is one of the classical graph neural network algorithms. It introduces the attention mechanism to aggregate the feature information of neighbors to extract the structural information of nodes. The training process of the GAT requires label information.
\item \textbf{DGI}: The deep graph infomax (DGI) is an unsupervised learning model that maximizes the common information between the target nodes and the global graph by establishing a contrastive perspective between the node level and the global graph level to generate the node embedding representations.
\item \textbf{GMI}: The graphical mutual information (GMI) is an unsupervised method that maximizes the mutual information of both features and edges between input graph and output graph of the encoder to learn node embeddings.
\item \textbf{SUBG-CON}: The sub-graph contrast (SUBG-CON) is a self-supervised learning paradigm, which maximizes the common information at the node-level and subgraph-level by constructing the subgraphs for nodes. The SUBG-CON method alleviates the memory problem of the GCL paradigm.
\item \textbf{GGD}: The graph group discrimination (GGD) is a self-supervised learning model, which trains the model by assigning different artificial labels to positive and negative samples to achieve the consistency of feature information between similar samples and then obtains the embedding representations of nodes.
\end{itemize}
% 实验设置
\subsection{Experimental Settings}
% 节点分类的参数设置
In the experiments, for the graph $\mathcal{G} = (\mathcal{V}, \mathcal{E})$, we employ the personalized pagerank (PPR) algorithm to obtain the top $K$ important node sets of target nodes for constructing the semantic subgraphs. We set $K=20$ on five datasets in the experiments. 
For the parameter $\alpha$ in the margin triplet loss function, we set $\alpha=0.5$ on the four datasets except that the value of $\alpha$ in the Pubmed dataset is 0.35. In the GNN encoder, the input dimension is the initial feature dimension of the nodes, and the output dimension is 1000, except for 450 in the Pubmed dataset.
We use the Adam optimizer with an initial learning rate of 0.001 to adjust the parameters of the encoder. The experiments are conducted on an eight-core Intel i7 2.50 GHz processor and 16 GB of RAM and a NVIDIA RTX 3090 24G GPUs. 
% 节点分类任务
\subsection{Node Classification}
Compared to supervised learning methods, the proposed LS-GCL model exploits structural feature information of nodes using known graph structures (e.g. adjacency matrices), without the node labels, to learn the node embeddings.
We then use a small number of labels, such as 20 labels for each category of nodes, to fine-tune\cite{wang2021self} the node embeddings for specific node classification task.

Experimental results of node classification in terms of accuracy on five real-world datasets are shown in Table \ref{tab:table2}. All experimental results are the mean and standard deviation of ten repeated node classification experiments.
We can find that the proposed LS-GCL model obtains optimal node classification accuracy on all five datasets.
Our LS-GCL model mines the semantic information of nodes from both local and global perspectives. Nodes with similar structural features in the graph generate similar node embeddings, which helps to predict the unknown labels of nodes.
In addition, the node classification accuracy of the GCL models is higher than that of the semi-supervised learning methods on five datasets, demonstrating that the GCL methods are useful for capturing latent structural features of nodes.

% 节点分类的实验结果
\begin{table}[htbp]
\renewcommand{\arraystretch}{1.3}
\setlength{\abovecaptionskip}{0pt}%    
\setlength{\belowcaptionskip}{10pt}%
\caption {Experimental results of node classification in terms of accuracy on five real-world datasets.}
\label{tab:table2}
\centering
\resizebox{\textwidth}{!}{%修改字体以适应表格太宽
\begin{tabular}{ccccccc}
\hline
Methods  & Available data & Cora                    & Citeseer                & Pubmed                  & Cora\_ML                & DBLP                    \\ \hline
GCN      & X, A, Y        & 81.5±0.3                & 70.3±0.5                & 78.8±0.2                & 82.4±0.8                & 75.4±0.5                \\
GAT      & X, A, Y        & 82.9±0.7                & 71.8±0.7                & 79.2±0.3                & 84.6±0.7                & 74.2±0.4                \\
DGI      & X, A           & 81.3±0.5                & 71.6±0.5                & 77.7±0.5                & 80.6±0.2                & 78.3±0.3                \\
GMI      & X, A           & 82.6±0.2                & 69.2±0.4                & 78.2±0.6                & 81.7±0.4                & 78.6±0.6                \\
SUBG-CON & X, A           & 83.5±0.5                & 72.7±0.6                & 79.9±0.4                & 84.3±0.1                & 80.3±0.5                \\
GGD      & X, A           & 83.8±0.4                & 72.8±0.8                & 80.9±0.5                & 83.1±0.9                & 80.8±0.3                \\
LS-GCL   & X, A           & {\textbf{\underline{84.4±0.4}}} & { \textbf{\underline{73.0±0.3}}} & {\textbf{\underline{81.5±0.6}}} & { \textbf{\underline{86.9±0.5}}} & {\textbf{\underline{80.9±0.4}}} \\ \hline
\end{tabular}}
\end{table}

\subsection{Link Prediction}

The task of link prediction is to predict potential node-pair relationships according to the structural features between nodes in the graph.
Traditional GNN models, which are the supervised learning methods, require a large number of labels to train the parameters of the model for capturing the relationships between node pairs for link prediction experiments\cite{yang2023graph}.
The LS-GCL learns target node embeddings in the link prediction task is similar with the node classification experiments.
Our model migrates the node embeddings obtained from pre-training process to the downstream tasks by fine-tuning.
The link embeddings are generated by concatenating the embedding vectors of the node pairs. The training and test sets containing positive and negative samples as well as the corresponding link labels are fed into a classifier (e.g. random forest\cite{rigatti2017random}) to evaluate the effectiveness of the node embeddings generated by the LS-GCL model. In the five datasets, the proportion of test set is 40\%. 

\begin{figure*}[!t]
\centering
\subfloat{\includegraphics[width=0.33333\linewidth]{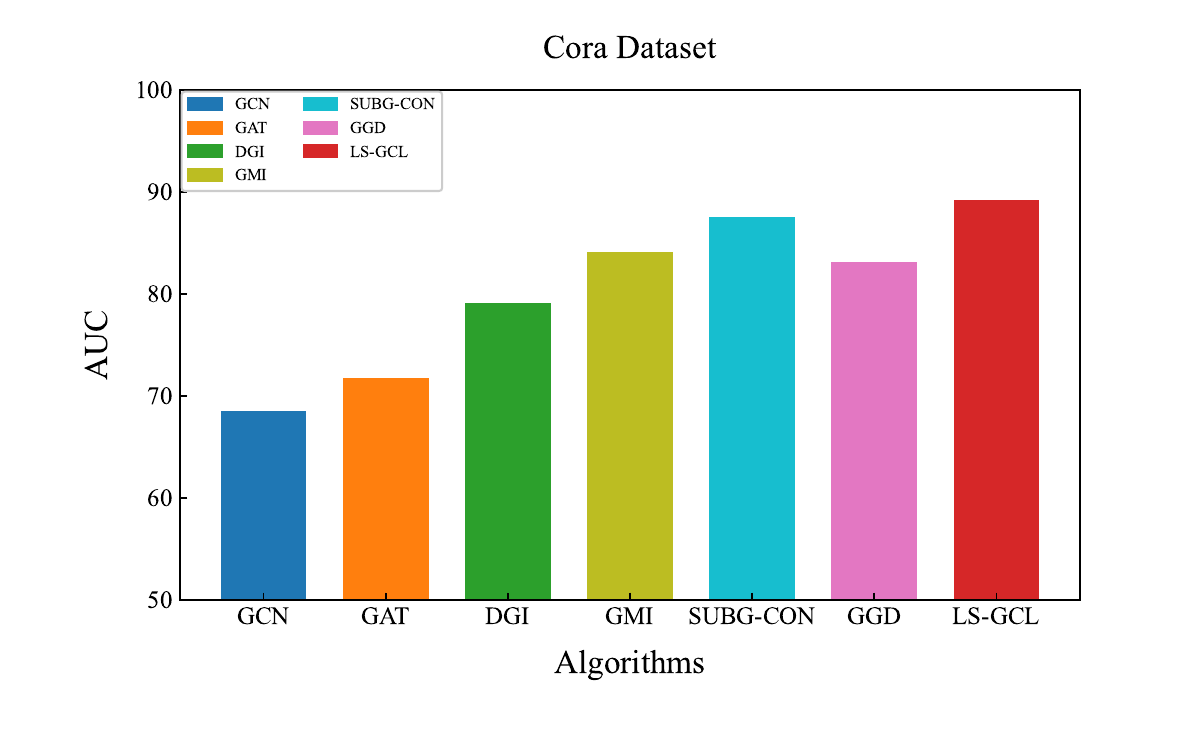}\label{1}}
% \hspace{-3mm}
\subfloat{\includegraphics[width=0.33333\linewidth]{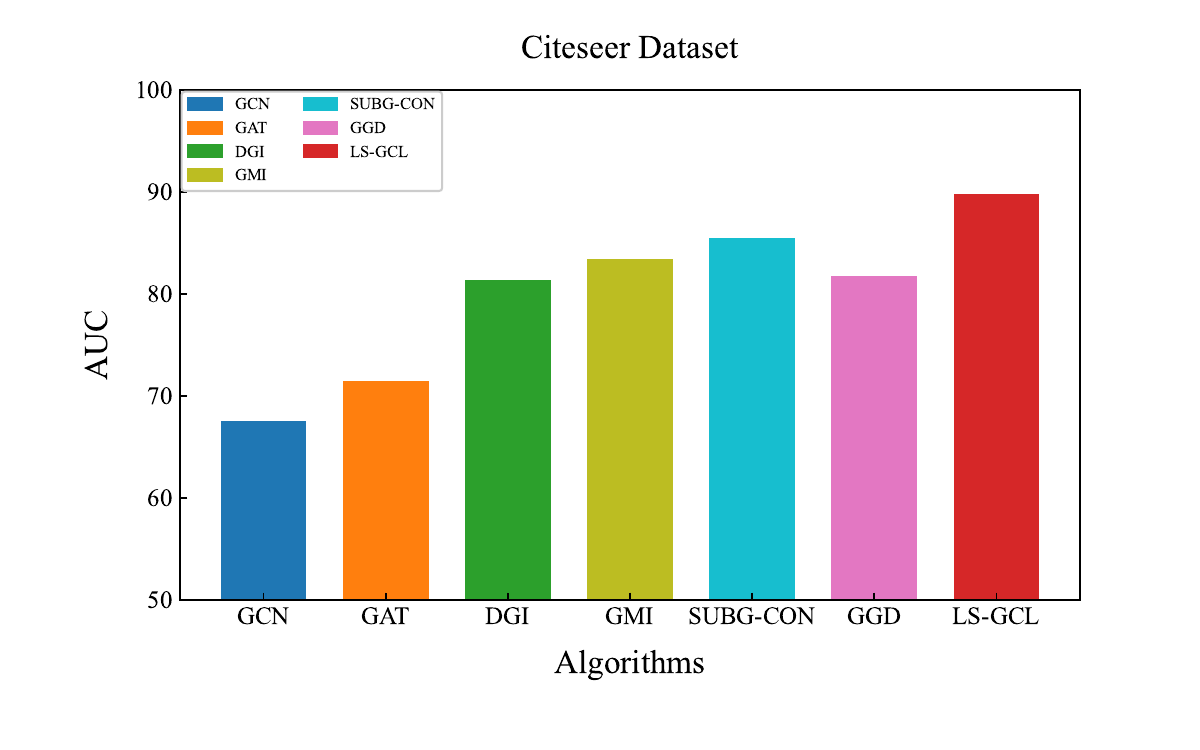}\label{2}}
% \hspace{-3mm}
\subfloat{\includegraphics[width=0.33333\linewidth]{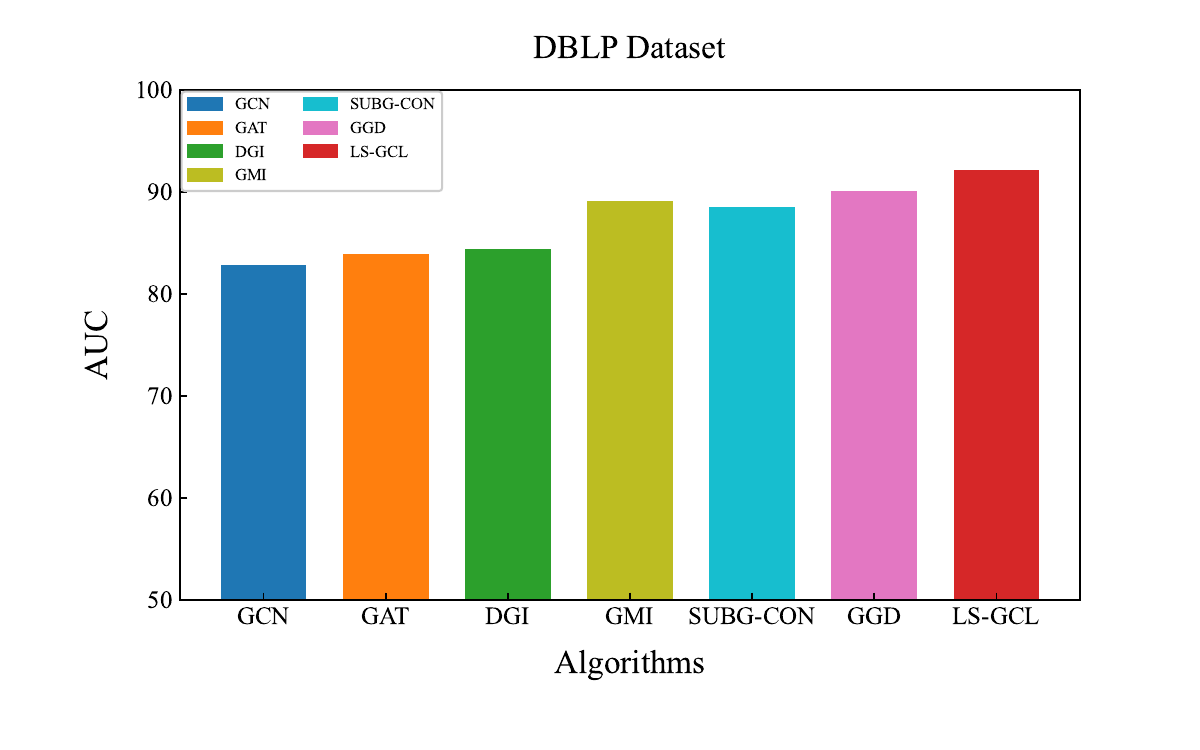}\label{3}}
% \hspace{mm}  % 调左右间隔
% \vspace{-4mm}  % 调上下间隔

\subfloat{\includegraphics[width=0.33333\linewidth]{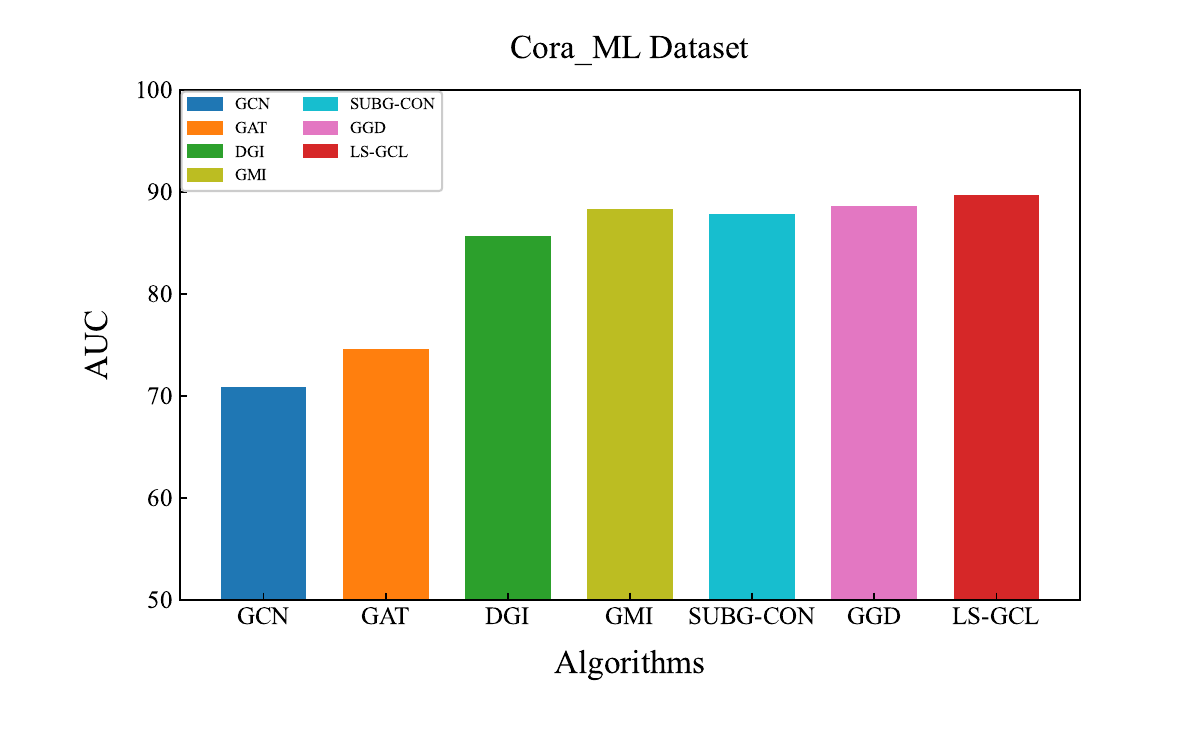}\label{4}}
% \hspace{-3mm}
\subfloat{\includegraphics[width=0.33333\linewidth]{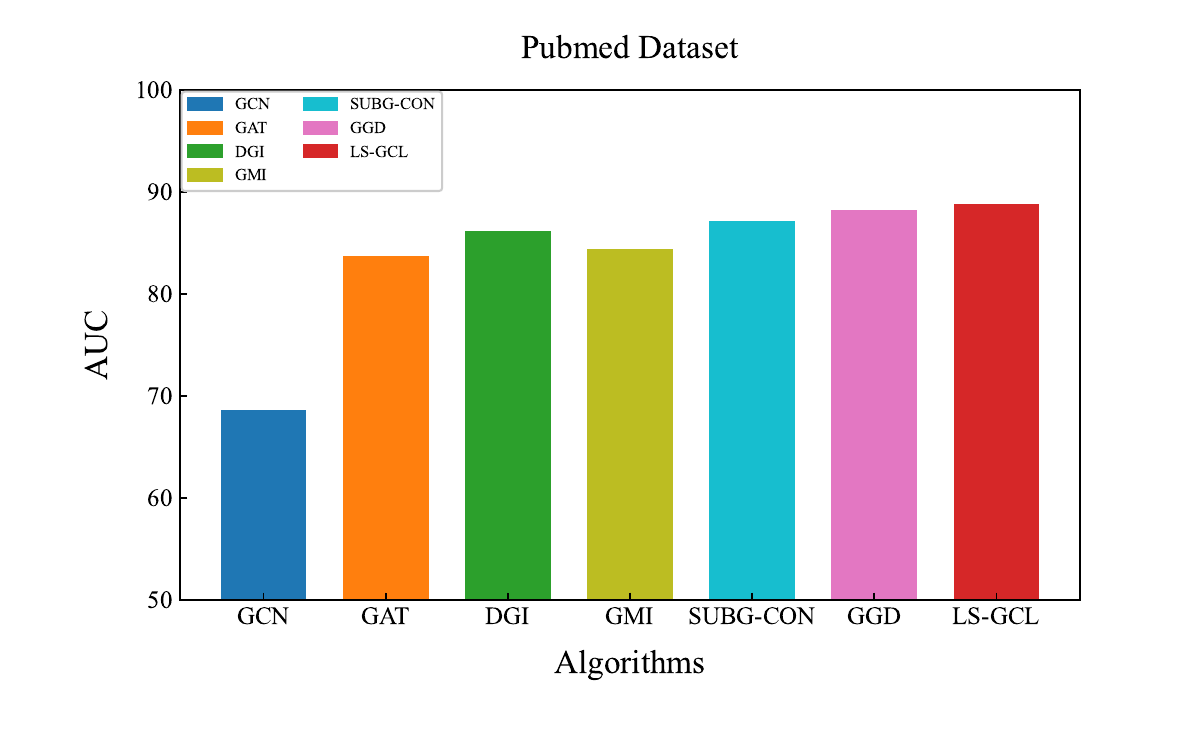}\label{5}}
% \hspace{-5mm}
\caption{The AUC of link prediction on five datasets for the proposed LS-GCL model.}
\label{fig:lp}

\end{figure*}

% 实验结果分析：
Experimental results of link prediction for the AUC index on five real-world datasets are shown in the Fig. \ref{fig:lp}. To illustrate the robustness of the proposed LS-GCL, we use three other metrics, namely Recall, Precision and F1-score(F1)\cite{yacouby2020probabilistic}, which are shown in Table \ref{tab:table3} and Table \ref{tab:table4}. All experimental results are the mean and standard deviation of ten repeated link prediction experiments.
We can see that our LS-GCL model achieves outstanding and stable experimental results on four evaluation indices for all five datasets compared to current GCL and semi-supervised learning models.
% The semantic subgraphs capture rich correlation information which is beneficial for analysing the characteristics of the links between pairs of nodes.
The node embeddings obtained by the GCL method with fine-tuning achieve an improvement in the link prediction task compared to the traditional GNN model.

% 链路预测的实验结果

\begin{table}[htb]
\renewcommand{\arraystretch}{1.3}
\setlength{\abovecaptionskip}{0pt}%    
\setlength{\belowcaptionskip}{10pt}%
\caption {Experimental results of link prediction for the Recall, Precision and F1 index on the Cora, Citeseer and Pubmed datasets.}
\label{tab:table3}
\centering
\resizebox{\textwidth}{!}{%修改字体以适应表格太宽
\begin{tabular}{cccccccccc}
\hline
Methods                       & \multicolumn{3}{c}{Cora}                                                       & \multicolumn{3}{c}{Citeseer}                                                   & \multicolumn{3}{c}{Pubmed}                                \\ \hline
\multicolumn{1}{c|}{}         & Recall            & Precision         & \multicolumn{1}{c|}{F1}                & Recall            & Precision         & \multicolumn{1}{c|}{F1}                & Recall            & Precision         & F1                \\
\multicolumn{1}{c|}{GCN}      & 56.9±0.2          & 74.2±0.4          & \multicolumn{1}{c|}{65.7±0.3}          & 52.8±0.3          & 75.6±0.2          & \multicolumn{1}{c|}{61.4±0.4}          & 64.4±0.3          & 70.3±0.1          & 67.2±0.1          \\
\multicolumn{1}{c|}{GAT}      & 50.4±0.2          & 88.7±0.3          & \multicolumn{1}{c|}{64.1±0.5}          & 51.9±0.5          & 87.2±0.6          & \multicolumn{1}{c|}{67.4±0.4}          & 72.5±0.4          & 85.3±0.5          & 81.4±0.3          \\
\multicolumn{1}{c|}{DGI}      & 73.0±0.6          & 83.1±0.8          & \multicolumn{1}{c|}{77.7±0.3}          & 69.6±0.5          & 91.1±0.3          & \multicolumn{1}{c|}{78.9±0.4}          & 84.0±0.9          & 88.1±0.9          & 85.9±0.6          \\
\multicolumn{1}{c|}{GMI}      & 74.1±0.3          & {\textbf{\underline{92.6±0.3}}} & \multicolumn{1}{c|}{82.3±0.2}          & 69.9±0.5          & {\textbf{\underline{95.8±0.1}}} & \multicolumn{1}{c|}{80.9±0.4}          & 80.6±0.2          & 87.3±0.7          & 83.8±0.6          \\
\multicolumn{1}{c|}{SUBG-CON} & 86.3±0.5          & 88.4±0.5          & \multicolumn{1}{c|}{87.3±0.4}          & 82.7±0.6          & 87.6±0.4          & \multicolumn{1}{c|}{85.1±0.4}          & 87.6±0.6          & 86.8±0.1          & 87.2±0.6          \\
\multicolumn{1}{c|}{GGD}      & 73.4±0.8          & 91.1±0.4          & \multicolumn{1}{c|}{81.3±0.6}          & 70.9±0.9          & 90.8±0.6          & \multicolumn{1}{c|}{79.6±0.5}          & 85.0±0.7          & {\textbf{\underline{90.2±0.9}}} & 87.8±0.5          \\
\multicolumn{1}{c|}{LS-GCL}   & \textbf{\underline{88.7±0.2}} & 91.6±0.4          & \multicolumn{1}{c|}{{\textbf{\underline{88.9±0.3}}}} & {\textbf{\underline{90.4±0.1}}} & 90.2±0.2          & \multicolumn{1}{c|}{{\textbf{88.8±0.4}}} & {\textbf{\underline{90.7±0.3}}} & 88.2±0.1          & {\textbf{\underline{88.3±0.4}}} \\ \hline
\end{tabular}}
\end{table}

% 链路预测的实验结果：

\begin{table}[htb]
\footnotesize
\renewcommand{\arraystretch}{1.3}
\setlength{\abovecaptionskip}{0pt}%    
\setlength{\belowcaptionskip}{10pt}%
\caption {Experimental results of link prediction for the Recall, Precision and F1 index on the Cora\_ML and DBLP datasets.}
\label{tab:table4}
\centering
\resizebox{\textwidth}{!}{%修改字体以适应表格太宽
\begin{tabular}{cllllll}
\hline
Methods                       & \multicolumn{3}{c}{Cora\_ML}                                                                        & \multicolumn{3}{c}{DBLP}                                                            \\ \hline
\multicolumn{1}{c|}{}         & \multicolumn{1}{c}{Recall} & \multicolumn{1}{c}{Precision} & \multicolumn{1}{c|}{F1}                & \multicolumn{1}{c}{Recall} & \multicolumn{1}{c}{Precision} & \multicolumn{1}{c}{F1} \\
\multicolumn{1}{c|}{GCN}      & 68.5±0.5                   & 72.5±0.4                      & \multicolumn{1}{l|}{69.1±0.2}          & 73.2±0.1                   & 88.7±0.6                      & 81.5±0.3               \\
\multicolumn{1}{c|}{GAT}      & 69.6±0.5                   & 79.1±0.3                      & \multicolumn{1}{l|}{72.8±0.5}          & 75.5±0.3                   & 91.7±0.2                      & 82.6±0.4               \\
\multicolumn{1}{c|}{DGI}      & 81.1±0.4                   & 89.4±0.4                      & \multicolumn{1}{l|}{85.0±0.3}          & 82.3±0.2                   & 86.0±0.7                      & 84.1±0.1               \\
\multicolumn{1}{c|}{GMI}      & 83.2±0.3                   & 92.6±0.2                      & \multicolumn{1}{l|}{87.7±0.1}          & 86.2±0.5                   & 91.4±0.2                      & 88.7±0.4               \\
\multicolumn{1}{c|}{SUBG-CON} & 84.4±0.4                   & 90.5±0.3                      & \multicolumn{1}{l|}{87.3±0.2}          & 89.6±0.1                   & 87.7±0.9                      & 88.6±0.9               \\
\multicolumn{1}{c|}{GGD}      & 84.5±0.3                   & 92.1±0.3                      & \multicolumn{1}{l|}{88.1±0.2}          & 87.9±0.1                   & 93.0±0.6                      & 90.9±0.1               \\
\multicolumn{1}{c|}{LS-GCL}   & \textbf{\underline{86.7±0.2}}          & \textbf{\underline{94.5±0.5}}                      & \multicolumn{1}{l|}{\textbf{\underline{89.4±0.3}}} & \textbf{\underline{93.1±0.2}}          & \textbf{\underline{93.4±0.1}}                      & \textbf{\underline{92.2±0.5}}      \\ \hline
\end{tabular}}
\end{table}

\section{Ablation Study}
% 参数分析
\subsection{Analysis of Subgraph Extraction Methods}

The LS-GCL model has achieved better results in both node classification and link prediction tasks.
In this section, we discuss three different ways of constructing subgraphs to evaluate the effectiveness of semantic subgraphs and the robustness of our LS-GCL framework.
$K$-hop subgraph\cite{duan2021residual} represents that the subgraphs are constructed by the neighboring nodes which are $K$ hop away from the target nodes. We construct the 1-hop subgraphs to capture the local structural features of target nodes. $K$-RW subgraph denotes that the local subgraphs are constructed by the path generated through the random walk algorithm with $K$ length for the target nodes. We set the length of the random walk as 20 to obtain the same size subgraph. $K$-rank subgraph denotes the semantic subgraph which contains top $K$ related nodes. 

Node classification results of three methods for extracting subgraphs are shown in Table \ref{tab:table5}. Our LS-GCL method with semantic subgraphs has higher accuracy than the LS-GCL model based on $K$-hop subgraphs and $K$-RW subgraphs, which indicates that the semantic subgraphs are capable for capturing potential structural information of nodes. The semantic subgraphs not only focus on the first-order neighboring nodes of target nodes, but also consider the higher-order related nodes, which enriches the semantic information of nodes.

\begin{table}[htb]
\footnotesize
\renewcommand{\arraystretch}{1.3}
\caption {Node classification accuracy of three methods for extracting subgraphs}
\label{tab:table5}
\centering
\begin{tabular}{llllll}
\hline
                & Cora              & Citeseer          & Pubmed            & Cora\_ML          & DBLP              \\ \hline
\textit{K}-hop   & 79.4±0.5          & 71.2±0.3          & 78.5±0.6          & 82.6±0.3          & 77.3±0.4          \\
\textit{K}-RW      & 79.1±0.6          & 71.1±0.4          & 80.2±0.5          & 81.6±0.7          & 76.1±0.8          \\
\textit{K}-rank  & \textbf{\underline{84.4±0.4}} & \textbf{\underline{73.0±0.3}} & \textbf{\underline{81.5±0.6}} & \textbf{\underline{86.9±0.5}} & \textbf{\underline{80.9±0.4}} \\ \hline
\end{tabular}
\end{table}

\subsection{Analysis of Subgraph Size}

\begin{figure}[htb]%强制固定位置
\centering
\includegraphics[width=0.7\linewidth]{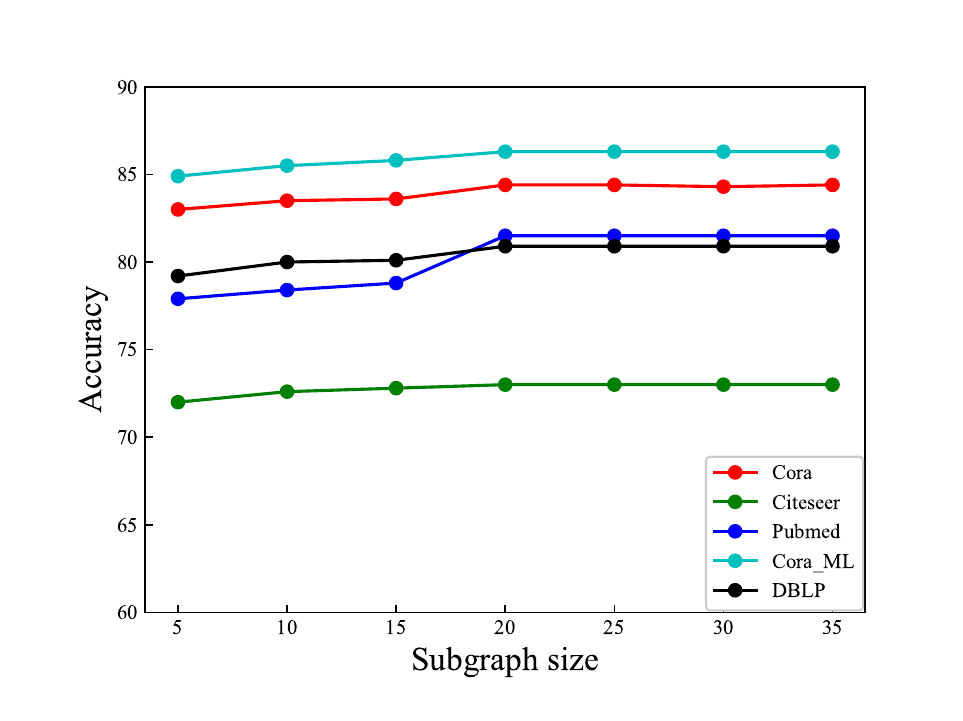}
\caption{The influence of subgraph size of node classification accuracy on five datasets.}
\label{fig:size}
\end{figure}

The semantic subgraphs of our LS-GCL model are constructed with the top $K$ relevant nodes. 
We investigate how subgraph size affects the actual performance of our model on node classification task, which is shown in the Fig.\ref{fig:size}. It has been shown that the node classification accuracy of the LS-GCL model steadily improves as the subgraph size $K$ increases.
For the five datasets, the accuracy gradually improves as the subgraph size $K$ increases from 0 to 20, indicating that GNN can extract more structural information from semantic subgraphs with more nodes.
The node classification accuracy of the LS-GCL model reaches a plateau when the number of nodes exceeds 20. 
Compared to traditional GCL methods such as the SUBG-CON, the LS-GCL model considers the global graph structure, which improves the robustness of our model in downstream tasks.

\subsection{Analysis of Encoder}

In this paper, the proposed LS-GCL uses the GNN model as an encoder to learn node embeddings.
We perform node classification experiments with different GNN models as feature encoders to test the impact of the encoders on our LS-GCL framework. For the GAT, the GraphSAGE\cite{hamilton2017inductive}, the SGC model\cite{wu2019simplifying}, we apply a single GNN layer as the encoders. The experimental results are shown in Table \ref{tab:table6}. 
Our LS-GCL framework using a GCN-based encoder achieves higher accuracy on four datasets except the DBLP dataset.
The node classification accuracy of the GNN-based encoders varies slightly between the five datasets, which demonstrates that our LS-GCL framework is robust. We can adapt the different GNN encoders to the actual situation and design a specific LS-GCL model for the downstream tasks.

\begin{table}[htb]
\footnotesize
\renewcommand{\arraystretch}{1.3}
\caption {Node classification accuracy for the four classic GNN encoders on node classification.}
\label{tab:table6}
\centering
\begin{tabular}{llllll}
\hline
          & Cora              & Citeseer          & pubmed            & Cora\_ML          & DBLP              \\ \hline
GCN       & \textbf{\underline{84.4±0.4}} & \textbf{\underline{73.0±0.3}} & \textbf{\underline{81.5±0.6}} & \textbf{\underline{86.9±0.5}}          & 80.9±0.4          \\
GAT       & 82.6±0.5          & 72.6±0.9          & 80.3±0.7          & 86.5±0.6          & \textbf{\underline{81.3±0.9}} \\
GraphSAGE & 84.0±0.5          & 72.1±0.7          & 79.4±0.2          & 86.1±0.4          & 79.1±0.2          \\
SGC       & 84.1±0.4          & 72.6±0.6          & 80.5±0.9          & 86.8±0.6 & 79.7±0.3          \\ \hline
\end{tabular}
\end{table}

\subsection{Analysis of Contrastive Loss Function}

\begin{figure}[htb]%强制固定位置
\centering
\includegraphics[width=0.7\linewidth]{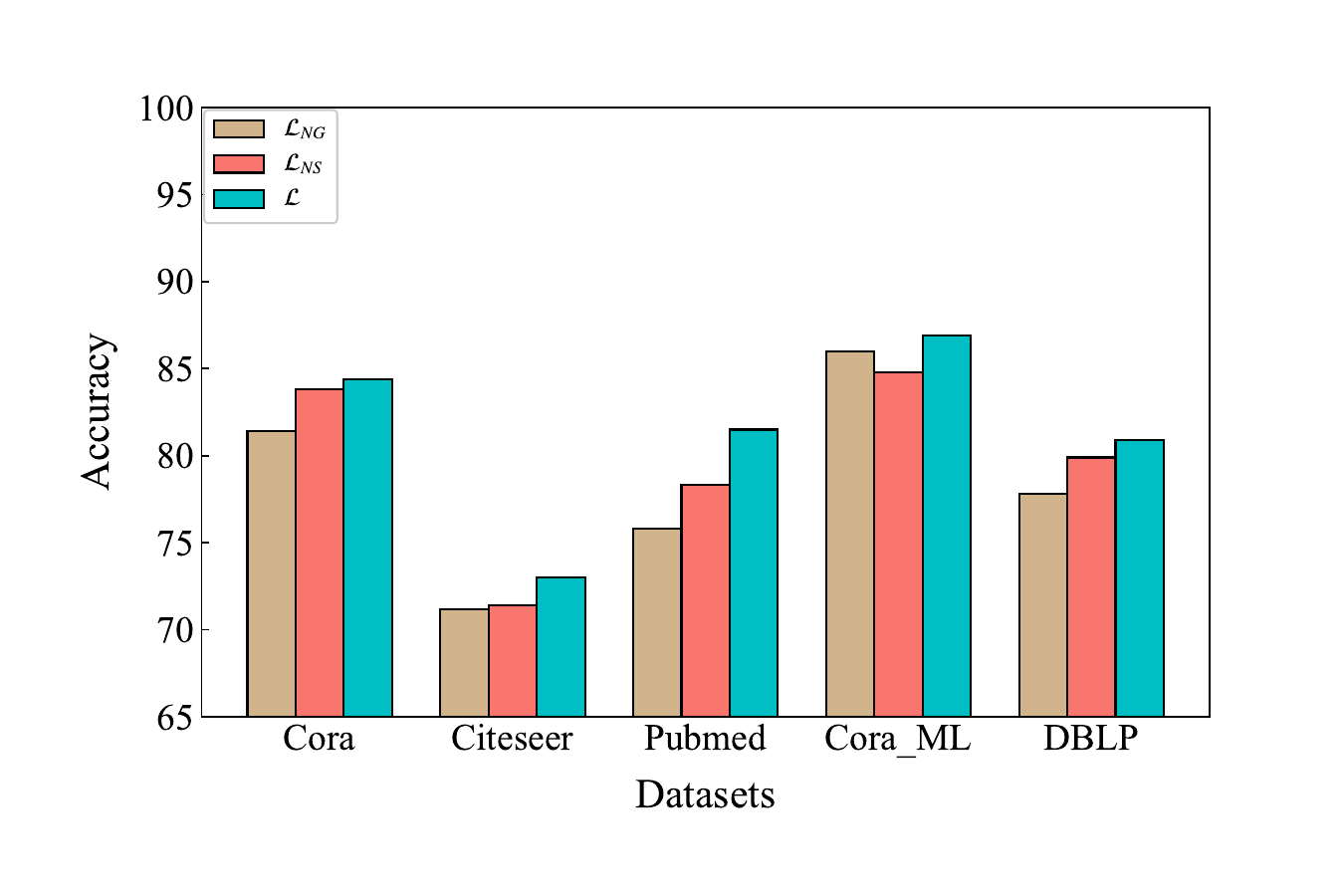}
\caption{Experimental results of node classification in terms of accuracy corresponding to different contrastive loss functions on five datasets.}
\label{fig:loss}
\end{figure}

In the paper, we propose a new multi-level contrastive loss function $\mathcal{L}$. To reflect the semantic information of nodes at different levels, we set the node embeddings at the global graph-level, the node embeddings at the subgraph-level and the subgraph-level graph embeddings corresponding to target nodes as the contrastive objectives.
To evaluate the effectiveness of the multi-level contrastive loss function, we use $\mathcal{L}_{NG}$ and  $\mathcal{L}_{NS}$ as the contrastive loss functions to learn the target node embeddings and perform node classification experiments, respectively. $\mathcal{L}_{NG}$ denotes that the contrastive objectives are node embeddings at subgraph-level and global graph-level.
$\mathcal{L}_{NS}$ denotes that the contrastive objectives are node embeddings at subgraph-level and subgraph-level graph embeddings corresponding to target nodes.
Experimental results of node classification experiments corresponding to different contrastive loss functions on five datasets are shown in Fig.\ref{fig:loss}.
The LS-GCL model using the multi-level contrastive loss function $\mathcal{L}$ achieves better results than $\mathcal{L}_{NG}$ and $\mathcal{L}_{NS}$.
% our $\mathcal{L}$ add local graph embeddings corresponding to target nodes, which captures the semantic information of the semantic subgraphs for target nodes.
% Compared to the contrastive loss function $\mathcal{L}_{NS}$, we consider the node embeddings at the global graph-level, which reduces the loss of structural information.
Compared with the models using $\mathcal{L}_{NG}$ and $\mathcal{L}_{NS}$ as the contrastive loss functions, the LS-GCL model captures different semantic information at three levels and each contrastive object has an essential role in the maximisation of the feature information for target nodes.

\section{Conclusion and Discussions}

In the paper, we propose a novel GCL framework named LS-GCL that pays attention to local structural information of nodes at multiple levels.
To capture more local structural information, we construct semantic subgraphs for nodes by the PPR algorithm. A shared GNN-based encoder is employed to mine the structural information of the local semantic subgraphs for target nodes. 
We obtain the target node embeddings at the subgraph-level and subgraph embeddings corresponding to target nodes to capture local structural features at different levels.
To preserve the potential semantic information of the original graph, we learn the target node embedding at the global graph-level.
Finally, we construct a multi-level contrastive loss function to maximize the common information of target nodes at different levels and obtain the final node embeddings.
We conduct node classification and link prediction experiments on various real-world datasets and achieve superior performance comparing with classical GNN and GCL models.

The proposed LS-GCL model requires numerous iterations to learn the parameters of encoder, which uses more running memory than traditional GNN models. How to alleviate the memory and time limitations of LS-GCL model is a public problem.
In addition, we aim to extend the LS-GCL framework to the heterogeneous information graphs that contain multiple types of nodes or relations\cite{zhu2022geometry}. The semantic subgraph proposed in our model will contribute to the mining of potential semantic information in heterogeneous graphs. In the future, the LS-GCL framework will be proposed to apply on the homogeneous graphs and heterogeneous graphs.

\section{Acknowledgments}
 This work is supported in part by the Natural Science Foundation of the Jiangsu Higher Education Institutions of China(No.22KJD120002).

\bibliographystyle{elsarticle-num} 
\bibliography{refs}

\end{document}